\documentclass{article}

\usepackage{iclr2027_conference,times}

\usepackage[utf8]{inputenc}
\usepackage[T1]{fontenc}
\usepackage{hyperref}
\usepackage{url}
\usepackage{booktabs}
\usepackage{tabularx}
\usepackage{amsfonts}
\usepackage{amsmath}
\usepackage{amssymb}
\usepackage{nicefrac}
\usepackage{microtype}
\usepackage{graphicx}
\usepackage{xcolor}
\usepackage{multirow}
\usepackage{caption}
\usepackage{subcaption}
\usepackage{tikz}
\usetikzlibrary{shapes, arrows.meta, positioning, fit, calc}
\usepackage{wrapfig}
\usepackage{colortbl}
\usepackage{placeins}
\usepackage{float}

\newcommand{\musr}{\textsc{MuSR}}
\newcommand{\dup}[1]{\textsubscript{\scriptsize\textcolor{green!50!black}{+#1}}}
\newcommand{\ddown}[1]{\textsubscript{\scriptsize\textcolor{red!75!black}{$-$#1}}}

\newcommand{\method}{ARISE}
\newcommand{\Method}{ARISE}

\iclrfinalcopy

\title{Learning Reusable Agent Subroutines\\from Experience}

\author{Zhihan Lei, Jiarui Yan, Joshua Momo \& William W. Cohen \\
Carnegie Mellon University}

\begin{document}

\maketitle
\lhead{}

\begin{abstract}
 Existing methods induce various forms of knowledge from agent traces, from sample workflows to programmatic skills. We propose \Method{} (Agent Routine Induction from Successful Experience), which identifies recurring procedures in agent traces and synthesizes them into a reusable library of agent subroutines. Across five tasks in narrative deduction, rule application, and constraint planning, it is shown to be a more effective learning strategy than prompting methods and prose and code artifact induction methods. \Method{} outperforms chain-of-thought in all ten model--task settings of the main evaluation. Compared to state-of-the-art methods in agent experience induction, our method scores 6.1 percentage points over the strongest evaluated baselines, and 11.5 percentage points over code-skill induction. In our controlled experiment, ARISE also improves average accuracy by up to 15.6 percentage points compared to formatting the induced knowledge as skills and by up to 19.1 points over rewriting it as code tools.
\end{abstract}

\section{Introduction}
\label{sec:intro}

\begin{figure}[!b]
\centering
\includegraphics[width=0.96\linewidth]{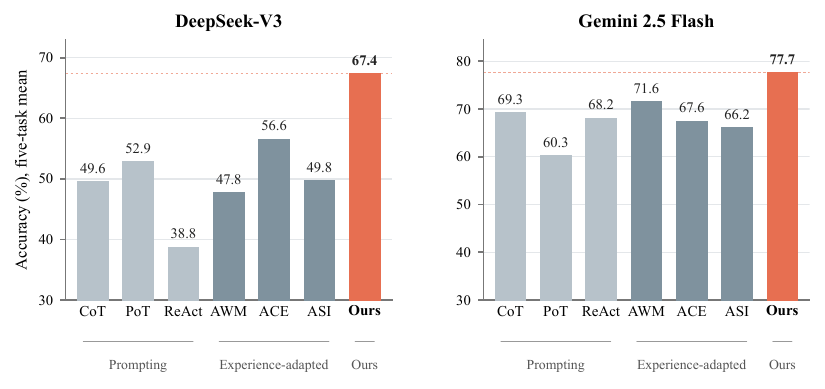}
\caption{\textbf{Test-set accuracy by model and method.} Each panel shows the unweighted mean of the five task columns of Table~\ref{tab:families}.}
\label{fig:hero}
\end{figure}

Solving complex problems with large language models (LLMs) often requires working through intermediate steps rather than generating an answer directly. How these steps are expressed and carried out is therefore an important part of reasoning-system design. Earlier methods like chain-of-thought prompting elicit intermediate reasoning in natural language, while Program-of-Thoughts expresses computations as programs executed by an external interpreter~\citep{wei2022chain,chen2023pot}. After the emergence of interactive agents, this organization extends to the
relationship between reasoning and action. For example, ReAct interleaves reasoning with actions and observations, enabling an agent to gather information and revise its approach as a task unfolds~\citep{yao2023react}, while Plan-and-Act separates high-level planning from the
execution of environment-specific actions~\citep{erdogan2025planandact}. Across these approaches, reasoning guides not only the final answer, but also which actions to take and how to use their outcomes to make progress.

Such agents typically begin with predefined tools and environment actions, while task-specific routines must either be supplied by developers or worked out during execution. Experience-based adaptation offers a way to acquire these routines automatically. Agent
Workflow Memory extracts reusable sample workflows from past trajectories, Agent Skill Induction constructs and verifies programmatic skills, and ACE accumulates procedural guidance in an evolving playbook in an agent's context~\citep{wang2024awm,wang2025asi,zhang2025ace}.
These approaches reduce the need to manually specify how an agent should handle recurring tasks.

\begin{figure}[t]
\centering
\includegraphics[width=0.92\linewidth]{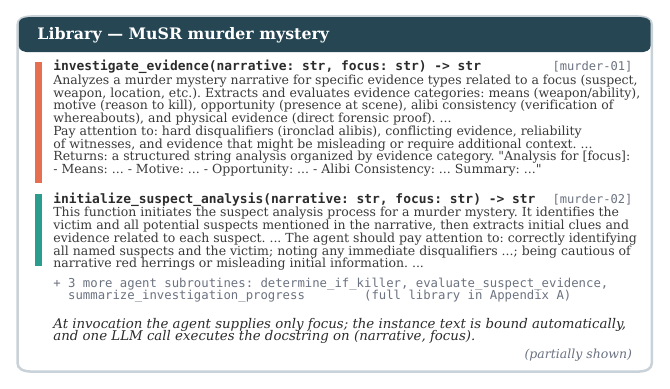}
\caption{\textbf{Agent subroutines induced on murder mystery solving for \musr{}.} Excerpts from two of the five subroutine specifications, showing their execution signatures. The complete library is described in Appendix~\ref{appendix:gallery}.} 
\label{fig:artifact}
\end{figure}

Yet, what an agent should learn from experience remains a question. We propose \Method{}, which learns a library of agent subroutines from successful agent trajectories. Each agent subroutine specifies a recurring operation in natural language and can be invoked by the agent similarly to a skill or a tool. This differs from prior work in that the learned subroutines are not put in an agent's context or executed by a deterministic program, but rather executed by an LLM that interprets the natural-language specification of the subroutine as shown in Figure~\ref{fig:artifact}. Our experiments isolate this execution choice: given the identical induced text, separate-call execution outperforms in-context skill retrieval in all ten model--task settings.

Agents repeatedly perform similar subroutines across different instances of a task. In murder-mystery problems, for example, an agent may reconstruct suspects' timelines, assess their motives, and compare evidence for
competing explanations~\citep{sprague2024musr}. In rule-guided reasoning, it must identify applicable rules and check whether their conditions hold in the current scenario~\citep{zhou2025rulearena}. Although the evidence and intermediate conclusions change, the
procedures for examining them often recur. These recurring subroutines suggest a useful level at which to reuse experience: not the answer to a previous problem or its entire solution trajectory, but a procedure that can be applied to new inputs. We extract these procedures into reusable subroutines.

We evaluate the approach on five tasks spanning narrative deduction, rule application, and constraint planning, with each model learning from its own executions and using the resulting library on held-out instances. Beyond overall accuracy, we examine how to execute learned agent subroutines, whether the learned libraries can transfer across models, and the extension to competition math problems. Our main contributions are:

\begin{itemize}
\item The execution interface determines the value of induced knowledge. With the induced text held fixed, agent subroutines outperform the same text delivered as retrievable skills in all ten model--task settings (+11.6 and +15.6 points on average on the two models) and the same procedures rewritten as Python tools in nine of ten (+7.9 and +19.1).
\item \Method{} performs up to 45.4 percentage points better than a ReAct agent, 6.1 to 10.8 points better than the strongest experience-adapted baseline on each model, and 11.5 points better than ASI (Figure~\ref{fig:hero}).
\item \Method{} does not require expert-designed tools: libraries induced with generic tools match those induced with hand-designed ones.
\item Agent subroutines induced by \Method{} are effective when transferred across models. Libraries induced by one LLM remain effective when deployed unchanged on another, improving accuracy over the target model's ReAct agent on all five tasks. 

\end{itemize}

\section{Related Work and Motivation}
\label{sec:related_work}

\paragraph{Reasoning structures and tools.} Prompting methods structure intermediate reasoning as chains, plans, trees, or programs~\citep{wei2022chain,wang2023planandsolve,yao2023tot,chen2023pot,gao2023pal}, and agents interleave such reasoning with tool calls, search, and self-feedback~\citep{yao2023react,zhou2024lats,madaan2023selfrefine,erdogan2025planandact}. Tools can also be learned or created by the model: Toolformer learns when to call APIs, and LLMs As Tool Makers, CREATOR, CRAFT, and TroVE build reusable Python toolsets~\citep{schick2023toolformer,cai2024latm,qian2023creator,yuan2024craft,wang2024trove}; code can also serve directly as an agent policy~\citep{liang2023codeaspolicies}. Tool creation yields programs, whose procedures must be expressible in code; in our knowledge-format comparison, rewriting induced procedures as Python tools lowers accuracy in nine of ten model--task settings relative to agent subroutines (Section~\ref{sec:delivery_form}). Modular approaches delegate subtasks to separate prompts or LLM calls: Decomposed Prompting routes subtasks to specialized handlers, meta-prompting has one LM conduct expert instances of itself, and DSPy composes LLM calls behind typed signatures whose prompts are optimized automatically~\citep{khot2023decomp,suzgun2024metaprompting,khattab2024dspy,opsahlong2024mipro,yuksekgonul2024textgrad}. In these systems, a developer defines the modules and how they compose, so the decomposition is fixed before deployment. Each agent subroutine is likewise an LLM call behind a fixed signature, but the subroutines and their specifications are induced from trajectories, and the agent composes them at run time.

\paragraph{Learning from agent experience.} Agents store experience as verbal reflections, extracted insights, manuals, or playbooks~\citep{shinn2023reflexion,zhao2024expel,chen2024automanual,zhang2025ace}, as workflows or skill documents~\citep{wang2024awm,ni2026trace2skill}, or as executable skills~\citep{wang2023voyager,zheng2025skillweaver,wang2025asi}; related methods reuse thought templates, cheatsheets, behaviors, or programs of thought distilled from prior problems~\citep{yang2024bot,suzgun2025dc,didolkar2025metacognitive,sarch2024ical}. In-context artifacts grow the agent's prompt as experience accumulates, and programmatic skills execute outside the context but only for procedures expressible as code. AWM also explores exposing workflows as callable actions, which replay fixed sequences of environment operations rather than invoking an LLM computation. \method{} delivers experience as procedures executed by separate LLM calls, and we compare with AWM, ACE, and ASI on the same training split.

\paragraph{Agent design and library learning.} AFlow, ADAS, and Self-Discover search over or compose reasoning and workflow structures~\citep{zhang2025aflow,hu2025adas,zhou2024selfdiscover}; we instead fix the interaction loop and learn the operations available within it. Library learning extracts reusable abstractions from solutions to prior tasks, as program abstractions~\citep{ellis2021dreamcoder,bowers2023stitch,grand2024lilo} or as natural-language reasoning abstractions trained with reinforcement learning~\citep{qu2025rlad}. Program libraries are verified by executing them on prior solutions, and RLAD trains its abstraction generator with reinforcement learning; \method{} induces its library from a single pass of successful rollouts without updating model parameters. We apply this idea to agent trajectories, with library entries interpreted by an LLM.

\section{Inducing Agent Subroutines}
\label{sec:method}

In Section~\ref{sec:formalization}, we describe the problem statement, then introduce the pipeline (Section~\ref{sec:pipeline}), and describe how induced agent subroutines are deployed and executed during inference (Section~\ref{sec:module_execution}).

\begin{figure}[t]
\centering
\includegraphics[width=\linewidth]{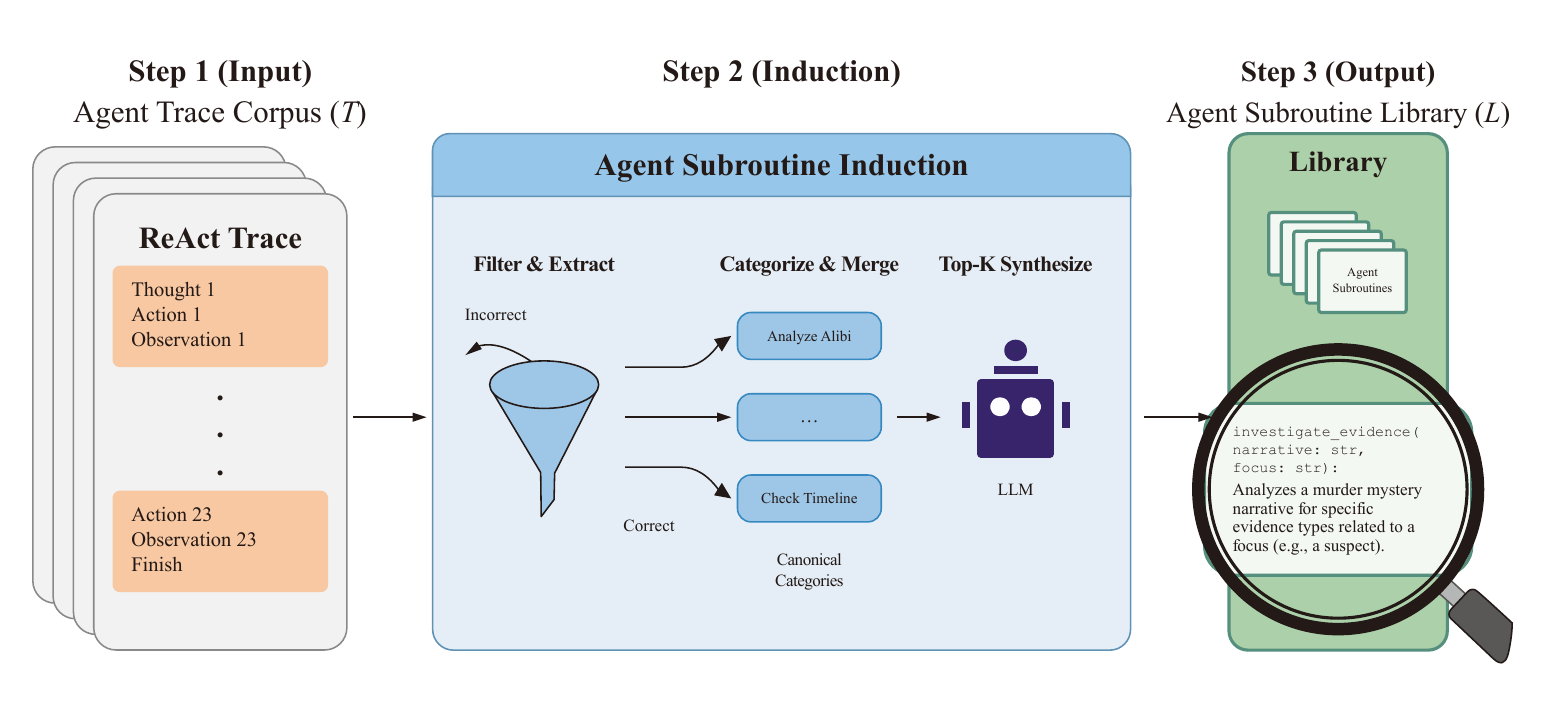}
\caption{Core Induction Pipeline as described in Section~\ref{sec:method}.}
\label{fig:pipeline}
\end{figure}

\subsection{Problem Formulation}
\label{sec:formalization}

Given a task family $\mathcal{F}$, we collect execution trajectories from a language-model agent on training instances and retain those with correct final answers, denoted by $\mathcal{T}^{+}$. We seek to construct a library of reusable agent subroutines from these trajectories for use on held-out instances of the same task family. The induction procedure maps
\[
\mathcal{T}^{+}\ \longmapsto\ \mathcal{L}=\{p_i\}_{i=1}^{M},
\qquad M\leq K,
\]
where $K$ is the maximum library size. Correctness is assessed at the trajectory level; individual intermediate steps are not independently verified.

We represent each agent subroutine with a calling interface and an execution specification:
\[
p_i=(n_i,\sigma_{\mathcal{F}},d_i),
\]
where $n_i$ is a name, $\sigma_{\mathcal{F}}$ is a predefined input--output signature shared within the task family, and $d_i$ is a natural-language specification of the subroutine. Induction determines the subroutines included in the library and produces their names and specifications. For example, a subroutine for narrative reasoning may assess the evidence concerning a particular suspect, taking the narrative and an agent-specified focus as inputs and returning a textual assessment.

\subsection{Induction Pipeline}
\label{sec:pipeline}

Let $\mathcal{T}=\{t_1,\dots,t_N\}$ be the rollouts of a ReAct agent on the training instances of a task family, with each rollout $t=((\textit{Th}_1,A_1,O_1),\dots,(\textit{Th}_L,A_L,O_L))$ representing an interleaved sequence of thoughts, actions, and observations. The agent is run once on the training split and its rollouts are analyzed once to induce a library $\mathcal{L}$ with $|\mathcal{L}|\leq K$. The procedure (Figure~\ref{fig:pipeline}; Appendix~\ref{appendix:impl}) has four stages.

\begin{enumerate}
\setlength{\itemsep}{1pt}\setlength{\parskip}{0pt}\setlength{\topsep}{2pt}
\item \emph{Filter.} Retain the rollouts whose final answer matches the label, $\mathcal{T}^+=\{t\in\mathcal{T}:\textsc{Correct}(t)\}$ (using task-specific correctness checks).
\item \emph{Extract.} Collect the per-step ReAct traces.
\item \emph{Categorize and merge.} One LLM call labels each item with a short reasoning-move name (e.g., ``verify alibi consistency''); if more than ten distinct labels result, a second call merges synonyms into 5--10 canonical categories.
\item \emph{Top-$K$ synthesis.} Categories are sorted by frequency. For each category with support at least $m$ (default $m{=}3$), the first five items in corpus order are shown to an LLM, which emits an agent subroutine name and docstring $(n,d)$; the signature $\sigma$ is fixed. Synthesis stops at $K$ agent subroutines or when no remaining category reaches $m$.
\end{enumerate}

Induction draws only on successful training trajectories, so it applies to tasks on which the base agent solves part of the training split. On Natural Plan trip and calendar, the base agent produces 2 and 19 successful trajectories; on trip, accuracy remains near zero for the induced libraries and every evaluated method (Table~\ref{tab:bootstrap_marginal}; Appendix~\ref{appendix:out_of_scope}).

\subsection{Subroutine Execution}
\label{sec:module_execution}
At inference, the induced agent subroutines are exposed as tools, replacing the base task-solving action set
with $\mathcal{L} \cup \{\textsc{finish}\}$.  Model parameters and the agent's underlying ReAct loop remain unchanged.
The agent sees subroutine names, argument schemas, and short descriptions, and selects a subroutine and
supplies its runtime arguments. The instance input is bound automatically. A separate LLM call
applies the subroutine's full specification $d_i$ to these inputs, without receiving the agent's interaction
history. The result is returned as an observation, which the agent incorporates into subsequent
reasoning and tool selection (Appendix~\ref{appendix:impl}).

The learned library determines which subroutines are available and what they compute; the agent
decides which subroutines to invoke and with what inputs on each instance. The library remains
fixed during evaluation, while the agent selects, repeats, and composes its subroutines as needed to solve
each new problem.

\FloatBarrier

\section{Experimental Setup}
\label{sec:setup}

\paragraph{Datasets.} Our main evaluation covers five tasks spanning narrative reasoning, rule application, and constraint planning. We additionally evaluate on competition mathematics to examine the method's applicability beyond these task families. We report accuracy on each task using the scoring criteria below. See Appendix~\ref{appendix:out_of_scope} for a discussion of tasks not included in the main evaluation.
\begin{itemize}
    \item \textbf{\musr{}}~\citep{sprague2024musr}. We evaluate three narrative-reasoning tasks: \emph{murder mystery}, identifying the culprit; \emph{object placement}, inferring where a character believes an object is located; and \emph{team allocation}, selecting an assignment of people to tasks under competing requirements. Instances are multiple choice, with two options on murder mystery, two to five on object placement, and three on team allocation; each is scored by exact match to the correct choice.
    \item \textbf{RuleArena NBA}~\citep{zhou2025rulearena}. This task evaluates rule-guided reasoning about proposed roster operations under the NBA Collective Bargaining Agreement. The relevant rule text is provided in the input. Each instance has a Boolean reference answer, and predicted compliance verdicts are scored by Boolean agreement.
    \item \textbf{Natural Plan: Meeting Planning}~\citep{zheng2024natural}. The agent constructs a meeting schedule subject to availability and travel-time constraints. A deterministic verifier checks schedule feasibility and counts the feasible meetings. A prediction is scored as correct when this count matches the reference schedule's.
    \item \textbf{AIME.} We evaluate on 43 problems from AIME 2023--2024; training trajectories are collected on 75 problems from AIME 1983--2020, so training and test problems do not overlap. Each problem requires an integer answer between 0 and 999 and is scored by exact match.
\end{itemize}

\paragraph{Baselines.} We compare against LLM prompting techniques (CoT and PoT), a ReAct agent (ReAct), and methods that enable agents to learn from experience (AWM, ACE, and ASI), which we call experience-adapted methods. Adaptation details are in Appendix~\ref{appendix:baselines}.
\begin{itemize}
    \item \textbf{Chain-of-Thought (CoT)}~\citep{wei2022chain} uses zero-shot prompting to generate a reasoning trace and answer in one model call, without external tools.
    \item \textbf{Program-of-Thoughts (PoT)}~\citep{chen2023pot} generates a Python program in one model call; sandboxed execution with standard-library support produces the answer.
    \item \textbf{ReAct}~\citep{yao2023react} interleaves reasoning with calls to the task's base tools. This unadapted agent also supplies the training trajectories for our induction procedure.
    \item \textbf{Agent Workflow Memory (AWM)}~\citep{wang2024awm} induces natural-language workflows from the same model and source training-rollout corpus as our method, and supplies them as contextual guidance.
    \item \textbf{Agentic Context Engineering (ACE)}~\citep{zhang2025ace} updates a natural-language playbook through generation, reflection, and curation during offline adaptation. The playbook is frozen and supplied as context during testing.
    \item \textbf{Agent Skill Induction (ASI)}~\citep{wang2025asi} induces executable Python skills from successful training interactions and verifies them through execution. Accepted skills become available during subsequent training interactions, and the final library is fixed during testing.
\end{itemize}

Both DeepSeek-V3 and Gemini 2.5 Flash are evaluated against all six baselines. Table~\ref{tab:adaptation_artifacts} summarizes the experimental configurations.

\begin{table}[t]
\centering
\caption{\textbf{Baseline methods compared in our experiments.} Settings, learned artifacts, and execution modes. Experience-adapted methods use ReAct as the base agent. Dashes indicate no learned artifact.}
\label{tab:adaptation_artifacts}
\small
\renewcommand{\arraystretch}{1.1}
\setlength{\tabcolsep}{4pt}
\begin{tabularx}{\linewidth}{@{}l >{\raggedright\arraybackslash}X l >{\raggedright\arraybackslash}X@{}}
\toprule
\textbf{Method} & \textbf{Setting} & \textbf{Learned artifact} & \textbf{Execution} \\
\midrule
CoT & Single LLM call & --- & Natural-language reasoning \\
PoT & Single LLM call & --- & Program execution \\
ReAct & Agent & --- & Reasoning and tool use \\
\midrule
AWM & Experience-adapted agent & Workflows & Contextual guidance \\
ACE & Experience-adapted agent & Playbook & Contextual guidance \\
ASI & Experience-adapted agent & Programmatic skills & Code execution \\
Ours & Experience-adapted agent & Agent-subroutine specifications & LLM execution \\
\bottomrule
\end{tabularx}
\end{table}

\paragraph{Models and evaluation protocol.} We use DeepSeek-V3 and Gemini 2.5 Flash for the main evaluation. We collect one ReAct rollout per training instance and induce our libraries from correct rollouts only. For the main part of our experiments, we use the same model to generate training trajectories, induce the library, and execute the resulting agent on test instances, without a stronger teacher model. The knowledge-format ablation uses both models; the base ReAct ablation uses Gemini 2.5 Flash; the rest of sensitivity sweeps use DeepSeek-V3. For AIME, we use Gemini 3.1 Flash-Lite and Gemini 2.5 Flash. To evaluate cross-model transfer, we deploy libraries induced by Gemini 2.5 Flash directly on Gemini 3.5 Flash-Lite without any modification.

\section{Results}
\label{sec:experiments}
\label{sec:main_results}
\label{sec:results}

\subsection{Main results}
\label{sec:families}

Table~\ref{tab:families} reports held-out accuracy across five tasks. As shown, induced libraries achieve the highest average accuracy among the evaluated methods on both models: 67.4\% on DeepSeek-V3 and 77.7\% on Gemini 2.5 Flash. These exceed the strongest baseline by five-task average on each model---ACE and AWM, respectively---by 10.8 and 6.1 percentage points.

\begin{table}[t]
\centering
\caption{\textbf{Main test-set results.} Test accuracy (\%). Bold marks the highest value per column within each model; Avg.\ is the unweighted five-task mean. Ours as skill and Ours as code deliver the same induced library as retrieved skill documents or Python tools (Section~\ref{sec:delivery_form}). $^\dagger$: served through a different provider (Appendix~\ref{appendix:baselines}).}
\label{tab:families}
\small
\renewcommand{\arraystretch}{1.05}
\setlength{\tabcolsep}{6pt}
\begin{tabular}{l l cccccc}
\toprule
& & \multicolumn{3}{c}{\musr{}} & RuleArena & NatPlan & Avg. \\
\cmidrule(lr){3-5}\cmidrule(lr){6-6}\cmidrule(lr){7-7}\cmidrule(lr){8-8}
\textbf{Method} & \textbf{Learned} & murder & object & team & NBA & meeting & \\
\midrule
\rowcolor{gray!12}\multicolumn{8}{l}{\textsc{DeepSeek-V3}} \\
\midrule
CoT & -- & 60.0 & 39.6 & 65.0 & 48.5 & 35.0 & 49.6 \\
PoT & -- & 67.0 & 60.4 & \textbf{69.0} & 40.9 & 27.0 & 52.9 \\
ReAct & -- & 45.0 & 50.0 & 38.0 & 28.8 & 32.0 & 38.8 \\
\midrule
AWM & sample workflows & 47.0 & 53.8 & 36.0 & 59.1 & 43.0 & 47.8 \\
ACE & context playbook & \textbf{77.0} & 50.0 & 58.0 & 62.1 & 36.0 & 56.6 \\
ASI & code skill & 71.0 & 50.0 & 55.0 & 47.0 & 26.0 & 49.8 \\
\midrule
Ours as skill$^\dagger$ & skill document & 67.0 & 47.2 & 64.0 & 43.9 & 37.0 & 51.8 \\
Ours as code$^\dagger$ & code tool & 66.0 & 57.5 & 65.0 & 34.8 & 18.0 & 48.3 \\
\rowcolor{gray!18}
\textbf{Ours (\method{})} & agent subroutine & 75.0 & \textbf{68.9} & 68.0 & \textbf{74.2} & \textbf{51.0} & \textbf{67.4} \\
\midrule
\rowcolor{gray!12}\multicolumn{8}{l}{\textsc{Gemini 2.5 Flash}} \\
\midrule
CoT & -- & 80.0 & 57.5 & 73.0 & 62.1 & 74.0 & 69.3 \\
PoT & -- & 72.0 & 72.6 & 60.0 & 56.1 & 41.0 & 60.3 \\
ReAct & -- & 78.0 & 50.9 & \textbf{76.0} & 65.2 & 71.0 & 68.2 \\
\midrule
AWM & sample workflows & 77.0 & 51.9 & 67.0 & \textbf{74.2} & \textbf{88.0} & 71.6 \\
ACE & context playbook & 80.0 & 72.6 & 70.0 & 65.2 & 50.0 & 67.6 \\
ASI & code skill & 75.0 & 62.3 & 59.0 & 69.7 & 65.0 & 66.2 \\
\midrule
Ours as skill & skill document & 78.0 & 50.9 & 71.0 & 66.7 & 64.0 & 66.1 \\
Ours as code & code tool & 82.0 & 61.3 & 68.0 & 72.7 & 65.0 & 69.8 \\
\rowcolor{gray!18}
\textbf{Ours (\method{})} & agent subroutine & \textbf{86.0} & \textbf{75.5} & \textbf{76.0} & 71.2 & 80.0 & \textbf{77.7} \\
\bottomrule
\end{tabular}
\end{table}

\begin{figure}[t]
\centering
\includegraphics[width=0.80\linewidth]{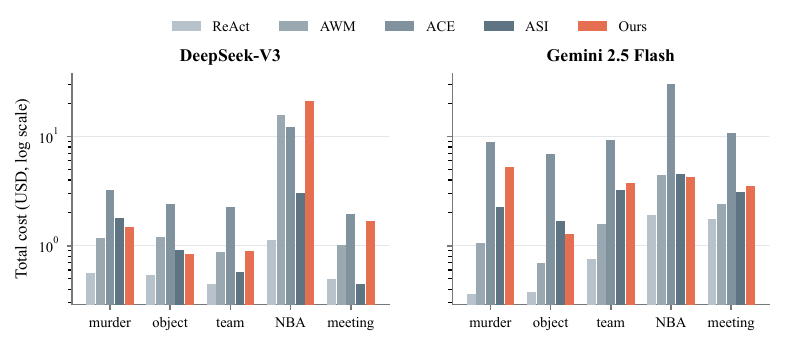}
\caption{\textbf{Total cost per task.} US dollars on a log scale for one offline adaptation plus test-set inference; values from Table~\ref{tab:compute_grid}.}
\label{fig:cost}
\end{figure}

Relative to the ReAct agents that supplied the trajectories, average accuracy rises from 38.8\% to 67.4\% on DeepSeek-V3 and from 68.2\% to 77.7\% on Gemini 2.5 Flash, and the induced libraries score at or above zero-shot CoT. On Gemini 2.5 Flash, \method{} outperforms ASI, which learns code artifacts, on all five tasks and by 11.5 percentage points on average. ACE costs 3.6$\times$ as much in total (\$13.14 vs.\ \$3.63 per task; Figure~\ref{fig:cost}) and scores 10.1 points lower; AWM is the cheapest experience-adapted method and scores 6.1 points lower. On DeepSeek-V3, the induced library has the highest average total cost in Table~\ref{tab:compute_grid} (\$5.23 per task), and NBA alone costs \$21.26. \method{} is the only experience-adapted method that matches or exceeds the ReAct agent in all ten model--task settings; on Gemini 2.5 Flash, the prose and code artifacts of ACE and ASI average below ReAct (67.6\% and 66.2\% vs.\ 68.2\%), while \method{} raises it to 77.7\%.

\subsection{Ablation and sensitivity analysis}
\label{sec:delivery_form}

\textbf{Knowledge format analysis.} The \emph{Ours as skill} and \emph{Ours as code} rows of Table~\ref{tab:families} hold each model's induction pipeline and library fixed and change only how the induced knowledge is delivered: as skill documents the agent retrieves by function call, or as Python tools rewritten from each specification (Appendix~\ref{appendix:delivery_formats}). The skill text is identical to the subroutine specification, so the gap between the \emph{Ours as skill} row and the \method{} row isolates executing a procedure in a separate LLM call from reading the same text in the agent's own context. Agent subroutines score higher than skills in all ten model--task settings and higher than code in nine; average accuracy rises from 51.8\% (skill) and 48.3\% (code) to 67.4\% on DeepSeek-V3, and from 66.1\% and 69.8\% to 77.7\% on Gemini 2.5 Flash, where code is 1.5 percentage points higher on NBA. On Gemini 2.5 Flash, delivered as skills, the induced text lowers accuracy below the ReAct agent that produced the trajectories (66.1\% vs.\ 68.2\%); delivered as agent subroutines, the same text raises it to 77.7\%. Whether induced knowledge helps at all depends on its execution interface.

\textbf{Content ablation.} Keeping the execution interface but removing the induced content has the opposite effect: replacing the specifications with a generic instruction, or additionally replacing the tool names, lowers the Gemini 2.5 Flash five-task average from 78.4\% to 67.7\% and 64.2\%, below the 68.2\% of the ReAct agent (Appendix~\ref{appendix:content_ablation}).

\begin{wraptable}{r}{0.42\linewidth}
\centering
\vspace{-1.2em}
\caption{\textbf{Induction under different trace-collection tool sets.} Test accuracy (\%) using libraries induced from successful trajectories collected with generic or hand-designed tools.}
\label{tab:trace_source}
\small
\setlength{\tabcolsep}{4pt}
\begin{tabular}{l cc}
\toprule
& \multicolumn{2}{c}{\musr{}} \\
\cmidrule(lr){2-3}
\textbf{Method / setting} & \textbf{Murder} & \textbf{Object} \\
\midrule
CoT                                              & 80.0 & 57.5 \\
ReAct                                          & 78.0 & 50.9 \\
\midrule
(a) generic                   & 86.0 & \textbf{75.5} \\
(b) hand-designed          & \textbf{87.0} & 70.8 \\
\bottomrule
\end{tabular}
\end{wraptable}
\textbf{Sensitivity to base ReAct settings.} We collect training trajectories under two tool settings for the base ReAct agent: (a) the generic text-search tools of the main table, which return matching sentences from the input, and (b) hand-designed tools written by experts for each task. At test time both settings replace their collection tools with the induced library. Libraries induced under both settings exceed CoT and ReAct on both tasks (Table~\ref{tab:trace_source}), and libraries induced with generic tools match those induced with expert-designed tools (86.0\% vs.\ 87.0\% on murder; 75.5\% vs.\ 70.8\% on object): \method{} requires no expert tool design.

\textbf{Sensitivity to hyperparameter choice.} Every maximum library size $K\in\{3,5,10\}$ exceeds the DeepSeek-V3 ReAct agent on all four tasks; $K{=}3$ and $K{=}5$ differ by at most 7.0 points per task, and $K{=}10$ is never best. Full discussion and detailed results are reported in Appendix~\ref{appendix:k_sweep}. 

\subsection{Extension to competition mathematics}
\label{sec:aime}

\begin{figure}[t]
\centering
\begin{minipage}[c]{0.60\linewidth}
\centering
\includegraphics[width=\linewidth]{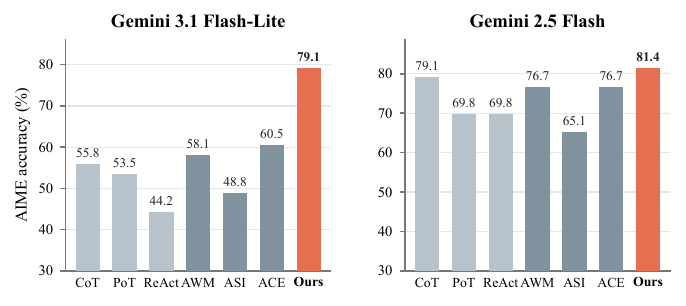}
\end{minipage}\hfill
\begin{minipage}[c]{0.37\linewidth}
\centering
\small
\setlength{\tabcolsep}{4pt}
\begin{tabular}{l cc}
\toprule
\textbf{Method} & \shortstack{3.1\\Flash-Lite} & \shortstack{2.5\\Flash} \\
\midrule
CoT & 55.8 & 79.1 \\
PoT & 53.5 & 69.8 \\
ReAct & 44.2 & 69.8 \\
AWM & 58.1 & 76.7 \\
ASI & 48.8 & 65.1 \\
ACE & 60.5 & 76.7 \\
\midrule
Ours (\method{}) & \textbf{79.1} & \textbf{81.4} \\
\bottomrule
\end{tabular}
\end{minipage}
\caption{\textbf{AIME accuracy by model and method.} Each panel plots one column of Table~\ref{tab:aime} (43 AIME 2023--2024 problems); within each model, AWM and the induced library reuse the same source training trajectories.}
\label{fig:aime}
\captionof{table}{\textbf{Competition mathematics results.} Accuracy (\%) of Gemini 3.1 Flash-Lite and Gemini 2.5 Flash on the same 43 AIME 2023--2024 problems. Each model induces its own library. Bold marks the highest accuracy per model.}
\label{tab:aime}
\end{figure}

We further evaluate the induction procedure on competition mathematics by learning a new library from successful AIME training trajectories. With Gemini 3.1 Flash-Lite, the induced library achieves 79.1\% accuracy, compared with 60.5\% for ACE, 58.1\% for AWM, 55.8\% for CoT, 53.5\% for PoT, 48.8\% for ASI, and 44.2\% for ReAct (Table~\ref{tab:aime}, Figure~\ref{fig:aime}). AWM and our method reuse the same source training-rollout corpus. Our method exceeds AWM by 21.0 percentage points.

With Gemini 2.5 Flash, a library induced separately from that model's successful trajectories achieves 81.4\% accuracy, compared with 79.1\% for CoT, 76.7\% for AWM and ACE, 69.8\% for both PoT and ReAct, and 65.1\% for ASI. The induced-library agent thus achieves the highest accuracy among the evaluated methods on both models, although the gain over the strongest baseline is smaller on Gemini 2.5 Flash. These results extend the evidence for learning reusable reasoning procedures to mathematical problem solving.

\subsection{Cross-model transfer}
\label{sec:transfer}

\begin{figure}[H]
\centering
\begin{minipage}[c]{0.50\linewidth}
\centering
\includegraphics[width=\linewidth]{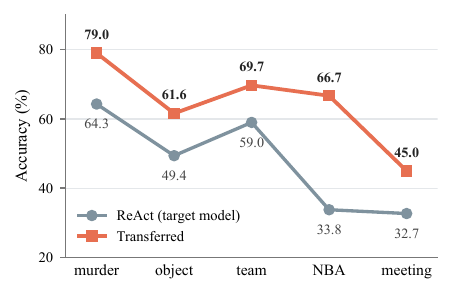}
\end{minipage}\hfill
\begin{minipage}[c]{0.47\linewidth}
\centering
\small
\setlength{\tabcolsep}{4pt}
\begin{tabular}{l c c r}
\toprule
\textbf{Task} & \textbf{ReAct} & \textbf{Transferred} & $\Delta$ (pp) \\
\midrule
murder  & 64.3 & \textbf{79.0} & $+14.7$ \\
object  & 49.4 & \textbf{61.6} & $+12.3$ \\
team    & 59.0 & \textbf{69.7} & $+10.7$ \\
NBA     & 33.8 & \textbf{66.7} & $+32.8$ \\
meeting & 32.7 & \textbf{45.0} & $+12.3$ \\
\bottomrule
\end{tabular}
\end{minipage}
\caption{\textbf{Cross-model transfer by task.} Gemini 3.5 Flash-Lite accuracy with its own ReAct agent and with libraries induced by Gemini 2.5 Flash; values from Table~\ref{tab:transfer}.}
\label{fig:transfer}
\captionof{table}{\textbf{Cross-model transfer of induced libraries.} Accuracy (\%) of Gemini 3.5 Flash-Lite using libraries induced by Gemini 2.5 Flash and deployed without modification, compared with the target model's ReAct agent. $\Delta$ denotes the paired accuracy gain in percentage points, computed from unrounded paired accuracies.}
\label{tab:transfer}
\end{figure}

 We deploy the libraries learned by Gemini 2.5 Flash on Gemini 3.5 Flash-Lite without modifying their content. Relative to Gemini 3.5 Flash-Lite's ReAct agent, the transferred libraries improve accuracy on all five tasks, with gains ranging from 10.7 percentage points on team allocation to 32.8 percentage points on NBA (Table~\ref{tab:transfer}, Figure~\ref{fig:transfer}). Cross-model transfer results show that the induced libraries are not model-specific, and can be used by other models without modification. 

\section{Discussion}
\label{sec:discussion}

Learning from agent experience has so far produced two kinds of artifacts: prose placed in the agent's context, such as workflows and playbooks, and code that the agent executes, such as programmatic skills. \method{} introduces a third: a natural-language procedure behind a callable interface, executed by its own LLM call. It keeps procedures in natural language, as prose methods do, and gives each one its own invocation, as code methods do. Across both models, it yields the highest five-task average among all evaluated methods and exceeds chain-of-thought in all ten model--task settings of the main evaluation. The gain comes from the execution interface. With the induced text held fixed, executing it as agent subroutines scores 11.6 and 15.6 percentage points above retrieving it as skills and 7.9 and 19.1 points above rewriting it as Python tools, on Gemini 2.5 Flash and DeepSeek-V3. The value of learned knowledge therefore depends on how it is executed as well as on what it says. The learned procedures do not depend on expert-designed tools: libraries induced with generic tools match those induced with hand-designed ones. Libraries transfer to Gemini 3.5 Flash-Lite without modification and improve it on all five tasks, and induced libraries give the highest AIME accuracy on both evaluated models. Agent subroutines thus provide a reusable form of experience that requires neither parameter updates nor expert tool design.

\bibliographystyle{plainnat}

\appendix
\noindent The appendices report the induced libraries, implementation and baseline details, ablations, cost, and statistical analyses.

\FloatBarrier
\section{Induced Agent Subroutines}
\label{appendix:gallery}

Below we present the induced agent subroutines evaluated in Table~\ref{tab:families} for both models. Induction is intended to capture common reasoning moves shared across instances of a task. Independently induced libraries exhibit recurring task-relevant agent subroutines despite differences in their names and granularity. A detailed comparison is in Appendix~\ref{appendix:skill_names}. These examples illustrate the level at which induction operates.

\FloatBarrier
\subsection{DeepSeek-V3 libraries}
\label{appendix:gallery_deepseek}

We summarize the DeepSeek-V3 libraries produced by the induction procedure in Section~\ref{sec:pipeline}, with a maximum size of $K{=}5$. The five main tasks have five agent subroutines each; the supplementary trip library evaluated in Table~\ref{tab:bootstrap_marginal} has two. The descriptions below summarize the agent subroutines; full specifications are retained in the library artifacts. Signatures are fixed by the synthesis prompt: \texttt{(narrative: str, focus: str)} for \musr{}, \texttt{(context: str, focus: str)} for RuleArena, \texttt{(prompt: str, focus: str)} for NatPlan; all return \texttt{str}.

\subsubsection{\musr{} murder}

\begin{itemize}
    \item \texttt{investigate\_evidence}: analyze evidence for a focused suspect, weapon, or location across categories means / motive / opportunity / alibi consistency / physical evidence; flag hard disqualifiers and conflicting witnesses.
    \item \texttt{initialize\_suspect\_analysis}: identify the victim and all suspects from the narrative; collect initial clues per suspect; output a structured JSON of victim, suspects, and per-suspect initial evidence.
    \item \texttt{determine\_if\_killer}: decide whether the focused suspect is the killer based on means, motive, opportunity, alibi consistency, and physical evidence; return \texttt{'1'} or \texttt{'0'}.
    \item \texttt{summarize\_investigation\_progress}: produce a structured per-suspect summary of case status with confidence levels, contradictions, and recommended next investigation steps.
    \item \texttt{evaluate\_suspect\_evidence}: weigh evidence for or against a focused suspect; produce a categorized list (motive / means / opportunity / physical evidence / alibi) and an overall assessment of evidence strength.
\end{itemize}

\subsubsection{\musr{} object placement}

\begin{itemize}
    \item \texttt{formulate\_search\_query}: extract entities and actions from the focus phrase to construct a precise search string targeting a specific narrative segment about object movement.
    \item \texttt{analyze\_character\_knowledge}: track when a character was present or absent during object movements; output the character's known locations, last-known location, and reasoning over knowledge gaps.
    \item \texttt{extract\_belief\_location}: determine where a target character believes a specific object is located, accounting for theory-of-mind divergence between actual location and the character's mental model.
    \item \texttt{infer\_character\_belief}: infer a character's belief about object location from observations and timing of arrivals/departures relative to movements; return believed location plus step-by-step reasoning.
    \item \texttt{track\_object\_movement\_timeline}: reconstruct a chronological timeline of object movements with witnesses and visibility annotations per event.
\end{itemize}

\subsubsection{\musr{} team allocation}

\begin{itemize}
    \item \texttt{analyze\_and\_summarize\_attributes}: extract per-person skills, weaknesses, and interpersonal dynamics from the narrative; bullet-point summary highlighting hard disqualifiers and tricky trade-offs.
    \item \texttt{seek\_specific\_individual\_information}: extract structured per-person profiles (strengths, weaknesses, characteristics, constraints) with attention to comparative language and conflicts.
    \item \texttt{define\_task\_requirements}: extract task list, required skills per task, hard/soft constraints, and explicit/implicit objectives; flag disqualifying conditions.
    \item \texttt{analyze\_team\_allocation}: produce structured analysis of people-tasks-constraints suitable for downstream allocation reasoning, including inferred skills and relative strength comparisons.
    \item \texttt{make\_final\_allocation\_decision}: select the best assignment option index given evaluated constraints; return integer index of the chosen option.
\end{itemize}

\subsubsection{RuleArena NBA}

\begin{itemize}
    \item \texttt{search\_cba\_rule}: locate relevant CBA rule text matching a focused term or section; return excerpt with section reference (e.g., ``Article XI, Section 4(b)'').
    \item \texttt{analyze\_operation\_compliance}: assess whether a proposed trade, signing, or offer satisfies CBA constraints given current salary cap and team status; output compliance status, violated rules, and conditions for compliance.
    \item \texttt{search\_cba\_term}: case-insensitive search for a defined CBA term, contract type, or threshold; return all matching context lines.
    \item \texttt{search\_cba\_section}: locate a specific section identifier (e.g., ``Section 8(e)(1)'') and return its full text with surrounding context.
    \item \texttt{search\_cba\_exception}: locate the named exception (e.g., Traded Player Exception, MLE) and return eligibility rules, calculation method, and stacking restrictions.
\end{itemize}

\subsubsection{NatPlan meeting}

\begin{itemize}
    \item \texttt{define\_meeting\_problem}: parse start location/time, friends with locations / availability windows / minimum durations, and the travel-time matrix; output structured JSON.
    \item \texttt{request\_travel\_times}: query the travel-time matrix for pairs involving a focused person or location; return a list of relevant travel durations.
    \item \texttt{propose\_meeting\_schedule}: synthesize a time-ordered schedule that maximizes friends met, respecting windows and travel; output sequence of arrival/departure events with travel segments.
    \item \texttt{verify\_constraints}: extract and validate constraints associated with a focused person or location; report constraint violations and missing information.
    \item \texttt{analyze\_constraints\_and\_requirements}: assess feasibility for the focused friend---earliest arrival, meeting window, minimum duration; output a feasibility verdict with timeline.
\end{itemize}

\subsubsection{NatPlan trip (\texorpdfstring{$|\mathcal{L}|{=}2$}{|L|=2})}

\begin{itemize}
    \item \texttt{identify\_constraints}: extract and analyze required cities, stay durations, total trip length, time windows, and flight-connectivity constraints; organize the findings by constraint type and flag possible conflicts.
    \item \texttt{analyze\_travel\_network}: analyze direct flight connections and potential routes in relation to required stays and time windows; summarize connectivity options and constraints affecting travel between cities.
\end{itemize}

\FloatBarrier
\subsection{Gemini 2.5 Flash libraries}
\label{appendix:gallery_gemini}

We summarize the Gemini 2.5 Flash libraries used for the five main tasks. Murder, object, and team have five agent subroutines each, NBA has four, and meeting has one. The maximum library size is $K{=}5$, not a requirement to produce exactly five agent subroutines. Signatures follow Appendix~\ref{appendix:gallery_deepseek}.

\subsubsection{\musr{} murder}

\begin{itemize}
    \item \texttt{investigate\_means\_and\_weapon}: identify the murder weapon and assess the focused suspect's access to or skill with it (or, for a focused weapon, the characters linked to it), noting when no link exists; return a JSON object keyed by weapon and means.
    \item \texttt{formulate\_killer\_conclusion}: name the most likely killer by weighing motive, means, opportunity, alibi consistency, physical evidence, and counter-evidence.
    \item \texttt{examine\_suspect\_profile}: extract a suspect's background, financial situation, relationships, history, and connection to the victim; return a JSON summary per category.
    \item \texttt{investigate\_motive\_and\_intent}: identify and evaluate motives (financial gain, revenge, secrets, power, jealousy, fear) and the nature of intent (premeditated, impulsive, accidental) for the focused person, with supporting snippets.
    \item \texttt{verify\_narrative\_facts}: search the narrative for a specific fact or detail; return \texttt{found}, matching snippets, and a summary message as JSON.
\end{itemize}

\subsubsection{\musr{} object placement}

\begin{itemize}
    \item \texttt{retrieve\_entity\_details}: return the sentences that explicitly mention the focused entity, without inference.
    \item \texttt{formulate\_character\_belief}: simulate the focused character's knowledge state over object movements, entrances, and exits; output ``The most likely place \{character\} would look to find the \{object\} is \{location\}.''
    \item \texttt{get\_character\_object\_belief}: track every object movement and the character's presence or awareness at each; deduce the believed location, which may differ from the actual one.
    \item \texttt{infer\_character\_beliefs}: track object movements and the character's presence, together with routines and roles that shape expectation; return object, character, believed location, and step-by-step reasoning.
    \item \texttt{identify\_character\_expected\_location}: return the last object location the character observed or was told of, with the focus given as ``character, object''; output a one-sentence answer.
\end{itemize}

\subsubsection{\musr{} team allocation}

\begin{itemize}
    \item \texttt{retrieve\_general\_information}: summarize everything in the narrative pertinent to the focus (a person's attributes or a task's requirements), flagging hard disqualifiers and implicit cues.
    \item \texttt{retrieve\_person\_profile\_details}: summarize the strengths, weaknesses, skills, experience, traits, and interpersonal dynamics of the focused people, related to task fit.
    \item \texttt{propose\_and\_finalize\_allocation}: propose an allocation of people to tasks and justify it against attributes, dynamics, task requirements, and constraints, including why alternatives are worse.
    \item \texttt{match\_skills\_to\_tasks}: find the textual evidence linking a person to the skill, task, or characteristic named in the focus.
    \item \texttt{retrieve\_relationship\_info}: return the exact sentences describing a specified relationship or interaction between individuals.
\end{itemize}

\subsubsection{RuleArena NBA (\texorpdfstring{$|\mathcal{L}|{=}4$}{|L|=4})}

\begin{itemize}
    \item \texttt{search\_cba\_rules}: return the verbatim CBA snippet matching the focused term (definitions, exceptions, thresholds).
    \item \texttt{explain\_nba\_cba\_violations}: identify each violation in the proposed operations, cite the CBA section, and give the rule and the calculation.
    \item \texttt{check\_nba\_cba\_compliance}: evaluate each proposed operation against cap and roster rules by operation type, player status, and team cap status; cite sections for compliant and non-compliant operations.
    \item \texttt{provide\_compliance\_verdict}: give the final verdict on the cumulative effect of all proposed operations on cap, luxury tax, roster limits, and exceptions.
\end{itemize}

\subsubsection{NatPlan meeting (\texorpdfstring{$|\mathcal{L}|{=}1$}{|L|=1})}

\begin{itemize}
    \item \texttt{generate\_meeting\_schedule}: from the start location and time, each friend's location, availability window, and minimum duration, and the travel-time matrix, produce a chronological step-by-step schedule that maximizes the number of friends met.
\end{itemize}

\FloatBarrier
\section{Comparing Libraries Across Inducing Models}
\label{appendix:skill_names}

Table~\ref{tab:skill_names} compares libraries independently induced by DeepSeek-V3 and Gemini 2.5 Flash from their own successful trajectories. Although the models produce different names and subroutine boundaries, their specifications recover related reasoning functions. The correspondence is semantic: agent subroutines address the same recurring demands of the task, rather than reproducing an identical list of tools.

\paragraph{Object placement.} Both models recover procedures for tracking character knowledge and inferring beliefs. DeepSeek-V3's \texttt{analyze\_character\_knowledge} tracks which object movements a character could observe, while \texttt{extract\_belief\_location} and \texttt{infer\_character\_belief} use that information to infer the character's believed location. Gemini's \texttt{get\_character\_object\_belief} and \texttt{formulate\_character\_belief} similarly reconstruct knowledge from movements, observations, and entrances or exits. Both libraries therefore capture the distinction between an object's actual location and the location supported by a particular character's information. DeepSeek also exposes timeline reconstruction as a separate agent subroutine; Gemini incorporates this reasoning into its belief-inference specifications.

\paragraph{NBA: retrieving rules and assessing compliance.} Both libraries contain agent subroutines for finding relevant CBA rules and evaluating proposed roster operations against them. DeepSeek separates retrieval by rule, term, section, and exception, alongside \texttt{analyze\_operation\_compliance}. Gemini uses a broader \texttt{search\_cba\_rules} agent subroutine and exposes compliance assessment, violation explanation, and a final verdict separately. The shared reasoning functions are rule retrieval and rule application; the models differ in where they place the boundaries between subroutines.

\paragraph{Murder: organizing and evaluating suspect evidence.} Both libraries organize the investigation around suspects and evidence concerning means, motive, opportunity, and alibis. DeepSeek's \texttt{investigate\_evidence} and \texttt{evaluate\_suspect\_evidence} combine several evidence categories within a call. Gemini exposes means and weapon analysis, motive analysis, and suspect profiling as separate agent subroutines, before forming a killer conclusion. The specifications thus recover related evidential reasoning while choosing different levels of specialization.

\paragraph{Shared functions, flexible granularity.} These correspondences do not require a fixed decomposition. For meeting planning, Gemini induces one schedule-generation subroutine, whereas DeepSeek distributes parsing, planning, and constraint assessment across several subroutines. The commonality lies in the task-relevant reasoning functions recovered, not a one-to-one match between subroutine names or counts. This comparison complements Section~\ref{sec:transfer}: independent induction reveals related procedural content across models, while cross-model execution demonstrates that a fixed library can remain useful beyond the model that produced it. Together, the two forms of evidence support interpreting the libraries as reusable representations of task-level reasoning.

\begin{table}[!htbp]
\centering
\caption{\textbf{Libraries independently induced by two models.} Each model uses its own successful trajectories. The inventories recover related task-level reasoning functions while differing in names and granularity; the accompanying analysis explains these semantic correspondences. These are the libraries evaluated in Table~\ref{tab:families}.}
\label{tab:skill_names}
\scriptsize
\setlength{\tabcolsep}{4pt}
\renewcommand{\arraystretch}{1.1}
\begin{tabular}{>{\raggedright\arraybackslash}p{0.08\textwidth} >{\raggedright\arraybackslash}p{0.42\textwidth} >{\raggedright\arraybackslash}p{0.42\textwidth}}
\toprule
\textbf{Task} & \textbf{DeepSeek-V3} & \textbf{Gemini 2.5 Flash} \\
\midrule
murder & \texttt{investigate\_evidence}, \texttt{initialize\_suspect\_analysis}, \texttt{determine\_if\_killer}, \texttt{summarize\_investigation\_progress}, \texttt{evaluate\_suspect\_evidence} & \texttt{investigate\_means\_and\_weapon}, \texttt{formulate\_killer\_conclusion}, \texttt{examine\_suspect\_profile}, \texttt{investigate\_motive\_and\_intent}, \texttt{verify\_narrative\_facts} \\
\addlinespace
object & \texttt{formulate\_search\_query}, \texttt{analyze\_character\_knowledge}, \texttt{extract\_belief\_location}, \texttt{infer\_character\_belief}, \texttt{track\_object\_movement\_timeline} & \texttt{retrieve\_entity\_details}, \texttt{formulate\_character\_belief}, \texttt{get\_character\_object\_belief}, \texttt{infer\_character\_beliefs}, \texttt{identify\_character\_expected\_location} \\
\addlinespace
team & \texttt{analyze\_and\_summarize\_attributes}, \texttt{seek\_specific\_individual\_information}, \texttt{define\_task\_requirements}, \texttt{analyze\_team\_allocation}, \texttt{make\_final\_allocation\_decision} & \texttt{retrieve\_general\_information}, \texttt{retrieve\_person\_profile\_details}, \texttt{propose\_and\_finalize\_allocation}, \texttt{match\_skills\_to\_tasks}, \texttt{retrieve\_relationship\_info} \\
\addlinespace
NBA & \texttt{search\_cba\_rule}, \texttt{search\_cba\_term}, \texttt{search\_cba\_section}, \texttt{search\_cba\_exception}, \texttt{analyze\_operation\_compliance} & \texttt{search\_cba\_rules}, \texttt{explain\_nba\_cba\_violations}, \texttt{check\_nba\_cba\_compliance}, \texttt{provide\_compliance\_verdict} \\
\addlinespace
meeting & \texttt{define\_meeting\_problem}, \texttt{request\_travel\_times}, \texttt{propose\_meeting\_schedule}, \texttt{verify\_constraints}, \texttt{analyze\_constraints\_and\_requirements} & \texttt{generate\_meeting\_schedule} \\
\bottomrule
\end{tabular}
\end{table}

\FloatBarrier
\section{Implementation Details}
\label{appendix:impl}

\paragraph{Scoring.} \musr{} uses exact answer-index agreement, NBA uses Boolean agreement, and AIME uses exact agreement of the extracted integer answer. Exceptions and missing answers count as incorrect. Meeting and trip results use a deterministic verifier to evaluate plans against their constraints. For meeting, the verifier simulates the parsed itinerary and compares its number of feasible meetings with that of the reference plan; for trip, it checks city durations, flight connections, and time constraints.

\FloatBarrier
\subsection{Induction and execution}

\paragraph{Induction prompts.} The fixed pipeline labels extracted items with short reasoning-operation names, merges synonymous labels, and synthesizes a named agent subroutine with a short description and full natural-language specification. Category support must be at least $m{=}3$; the default maximum library size is $K{=}5$. Synthesis uses the first five items of each category in corpus order. The same stages are used across benchmarks, with task descriptions and input argument names supplied as prompt parameters. Appendix~\ref{appendix:variants} reports the trace-filter ablation and the ablation of full specifications in the agent's prompt.

\paragraph{Deployment.} We replace the base action set with the induced library and \textsc{finish}. The agent sees agent subroutine names, argument schemas, and short descriptions. A separate LLM call executes each full specification on the runtime arguments and bound instance input, without receiving the agent's interaction history. Its output becomes an observation in that history. The library is fixed during test evaluation; the agent chooses when to call, repeat, and combine agent subroutines.

\paragraph{Decoding and request budgets.} The Gemini main-grid repeat evaluations, the ACE and ASI baselines, knowledge-format ablations, and AIME evaluations use $T{=}0$. The default ReAct limit is 50 model requests, with 200 for the Gemini 2.5 Flash NBA agent configurations, including ReAct, AWM, ASI, ACE, and the induced-library agent. These are limits on the outer agent's requests, not on tool invocations or all nested LLM calls. AIME's direct and agent subroutine execution calls allow up to 65\,536 output tokens. PoT generates a program in one model call; the Gemini 2.5 Flash NBA PoT evaluation uses a 600-second call timeout.

\FloatBarrier
\subsection{Knowledge format comparisons}
\label{appendix:delivery_formats}
In this experiment, the base ReAct agent is the same across three settings (skill, code and agent subroutines). The prompt and the harness are held fixed. Only the format of the induced knowledge differs. The agent sees the same short names and argument schemas in all three settings, but the full specifications are delivered in different formats.

\paragraph{Skill.} The agent retrieves the induced skills through function calls. Each skill's text is identical to the corresponding agent subroutine's specification. 

\paragraph{Code.} An agent rewrote each subroutine specification as a standard-library Python function that performs regex and keyword matching over the task input; calling the tool runs this function. A function was accepted only if it ran without errors on three training instances and produced varied outputs; brittle functions and functions that failed to generalize were rejected.

\FloatBarrier
\section{Baseline Adaptation}
\label{appendix:baselines}

All three experience-adapted baselines share the setting of \method{}: the same base ReAct agent and base tools, the same training split, the same model for adaptation and testing, and the same test split and request limits. Each learned artifact is frozen after adaptation. All three use gold labels during adaptation, as \method{} does through its correctness filter; ACE and ASI receive them as feedback on every update.

\paragraph{AWM.} We induce one workflow per task with a single LLM call over up to 20 successful training rollouts of the base ReAct agent, with each step truncated to its thought and action. The prompt asks for a numbered workflow of 5--10 steps. At test time the workflow is prepended to the agent's prompt as a learned guide.

\paragraph{ACE.} ACE adapts a playbook through generator, reflector, and curator roles~\citep{zhang2025ace}. The generator is the base ReAct agent with the current playbook; the reflector analyzes each trajectory together with the gold answer; the curator adds up to three bullets per problem, merged deterministically with helpful/harmful counters, near-duplicate removal, and a 10K-token budget. Adaptation runs five epochs over the training split (375 updates; 350 for meeting), five passes compared with the single pass of rollouts that \method{} analyzes, and yields playbooks of 88--266 bullets. The final playbook is supplied in the agent's prompt at test time.

\paragraph{ASI.} ASI makes one online pass over the training split~\citep{wang2025asi}. After each correct episode, one LLM call induces up to three Python skills and rewrites the trajectory with them; a skill is admitted only if re-solving the task with it reaches the correct answer, calls a newly induced skill, and raises no skill error. Episode correctness is judged against the gold label. Skills run in a sandbox and are available as tools in later episodes; final libraries contain 1--27 skills. At test time the skills are added to the agent's base tools.

\paragraph{DeepSeek-V3 serving.} The DeepSeek-V3 ACE, ASI, skill-delivery, and code-delivery runs use DeepSeek-V3 served through OpenRouter, as our original provider no longer serves the model; all other DeepSeek-V3 rows use the original provider.

\FloatBarrier
\section{Induction Under Different Base ReAct Settings}
\label{appendix:trace_source}
\label{appendix:handdesigned}
\label{appendix:trace_collection}

Table~\ref{tab:trace_source} provides results for induction under different base ReAct settings. Here we provide more details on the hand-designed tools and fine-grained results.

\FloatBarrier
\subsection{Hand-designed collection tools}
\label{appendix:handdesigned_tools}

The hand-designed condition uses the following three agent subroutines per task. They were authored for this study and executed through the same LLM execution interface as the induced agent subroutines. They are available only during trace collection in this ablation and are not added to its test-time libraries.

\paragraph{\musr{} murder.}
\begin{itemize}
    \item \texttt{extract\_suspects\_and\_evidence}: extract victim, crime details, and per-suspect evidence (motive, means, opportunity, alibi claim, alibi witnesses, suspicious behavior, physical evidence).
    \item \texttt{verify\_alibis}: cross-reference each suspect's alibi against the narrative; identify alibi gaps, contradictions, and corroborating evidence.
    \item \texttt{deduce\_murderer}: synthesize evidence and verified alibis to identify the perpetrator; weight physical evidence and alibi contradictions over motive alone.
\end{itemize}

\paragraph{\musr{} object placement.}
\begin{itemize}
    \item \texttt{extract\_movements}: enumerate every object-movement event chronologically with actor, source, destination, and lists of present/absent witnesses.
    \item \texttt{extract\_discoveries}: list incidental observations where a character learns an object's location without witnessing the move (saw directly, told by, inferred from event).
    \item \texttt{infer\_belief}: combine movements, discoveries, and a re-read of the narrative to determine where the target person believes the object is located.
\end{itemize}

\FloatBarrier
\subsection{Extracted items across collection settings}

Table~\ref{tab:trace_source_items} reports the items extracted from successful trajectories in the two collection settings above. A thought item is an extracted reasoning step; an action item is a tool-call record. These are item counts, not trajectory counts. The hand-designed murder condition supplies action items only.

\begin{table}[!htbp]
\centering
\caption{\textbf{Items extracted from successful collection trajectories.} Each condition uses the same 75 training instances per task. Counts distinguish thought items from action items supplied to induction; corresponding test accuracies appear in Table~\ref{tab:trace_source}.}
\label{tab:trace_source_items}
\small
\setlength{\tabcolsep}{7pt}
\begin{tabular}{l rrrr}
\toprule
& \multicolumn{2}{c}{\musr{} murder} & \multicolumn{2}{c}{\musr{} object} \\
\cmidrule(lr){2-3}\cmidrule(lr){4-5}
\textbf{Collection setting} & Thought items & Action items & Thought items & Action items \\
\midrule
(a) generic       & 23 & 91  & 31 & 72 \\
(b) hand-designed & 0  & 154 & 40 & 101 \\
\bottomrule
\end{tabular}
\end{table}
\FloatBarrier
\section{Content Ablation}
\label{appendix:content_ablation}

This ablation asks whether the gains of \method{} come from the content of the induced procedures or from executing each tool call in a separate LLM call. Table~\ref{tab:content_ablation} compares three configurations on Gemini 2.5 Flash. All three use the same base ReAct agent, the same number of tools, and the same execution interface: every tool call is executed by a separate LLM call on the instance input and the agent-supplied \texttt{focus}. Only the information attached to the tools differs.

\paragraph{ARISE.} The induced library as deployed in the main experiments: induced tool names and short descriptions in the agent's tool schema, and full natural-language specifications given to the executing LLM call.

\paragraph{ARISE with generic content.} Tool names, signatures, and argument descriptions are kept; every short description and full specification is replaced by the same generic instruction (``Helper reasoning tool. Provide a brief reasoning chain (2--4 sentences) identifying how you applied this approach, then give your final answer.'').

\paragraph{Generic tools.} Tools are renamed \texttt{helper\_1}, \ldots, \texttt{helper\_$k$} and carry no descriptions or specifications: the agent sees only the tool names and the \texttt{focus} argument, and the executing LLM call receives only the function signature.

\paragraph{Results.} The induced content accounts for most of the gain. Replacing the specifications with a generic instruction lowers the five-task average from 78.4\% to 67.7\%, and removing the induced names as well lowers it to 64.2\%. The execution interface alone does not reproduce the gains of \method{}: with generic tools, the five-task average falls below the 68.2\% of the ReAct agent in Table~\ref{tab:families}, and accuracy on meeting drops from 83.0\% to 44.0\%. The effect of content is concentrated on meeting and object, where \method{} exceeds generic tools by 39.0 and 21.7 percentage points, and it is also visible on team (+4.0 over generic tools, +9.0 over generic content). On murder and NBA, the three configurations lie within 4.5 points of one another; on these tasks, generic tools still exceed the ReAct agent of Table~\ref{tab:families} (84.0\% vs.\ 78.0\% on murder, 68.2\% vs.\ 65.2\% on NBA). Induced names carry part of the content: keeping them without specifications adds 3.5 points on average over generic tools, most of it on meeting (+15.0), whose library consists of a single tool.

\begin{table}[!htbp]
\centering
\caption{\textbf{Content ablation.} Test accuracy (\%) with Gemini 2.5 Flash. Subscripts give the difference from \emph{generic tools} in percentage points. ARISE values are from the evaluation runs paired with this ablation.}
\label{tab:content_ablation}
\small
\setlength{\tabcolsep}{8pt}
\begin{tabular}{l ccc}
\toprule
\textbf{Task} & \textbf{ARISE} & \textbf{ARISE with generic content} & \textcolor{gray}{\textbf{Generic tools}} \\
\midrule
\musr{} murder   & 87.0\dup{3.0}  & 83.0\ddown{1.0} & \textcolor{gray}{84.0} \\
\musr{} object   & 73.6\dup{21.7} & 55.7\dup{3.8}   & \textcolor{gray}{51.9} \\
\musr{} team     & 77.0\dup{4.0}  & 68.0\ddown{5.0} & \textcolor{gray}{73.0} \\
RuleArena NBA    & 71.2\dup{3.0}  & 72.7\dup{4.5}   & \textcolor{gray}{68.2} \\
NatPlan meeting  & 83.0\dup{39.0} & 59.0\dup{15.0}  & \textcolor{gray}{44.0} \\
\midrule
Avg.             & \textbf{78.4}\dup{14.2} & 67.7\dup{3.5} & \textcolor{gray}{64.2} \\
\bottomrule
\end{tabular}
\end{table}

\FloatBarrier
\section{Sensitivity Sweep}
\label{appendix:k_sweep}
\label{appendix:variants}

A useful library must provide enough specialized agent subroutines without making every possible instruction part of the agent's decision process. We examine this trade-off through library size and specification visibility, using DeepSeek-V3.

\paragraph{Library size.} Categories are processed in decreasing support until $K$ agent subroutines have been synthesized or no eligible categories remain; categories below $m{=}3$ are excluded and unsuccessful synthesis attempts are skipped. Table~\ref{tab:k_sweep} varies $K\in\{3,5,10\}$. A larger inventory is not consistently better: $K{=}3$ performs best on murder and object, while $K{=}5$ performs best on team and meeting. The $K{=}10$ library is never best in this sweep. These results support a compact library; we use $K{=}5$ as a common setting across tasks.

\begin{table}[!htbp]
\centering
\caption{\textbf{Sensitivity to maximum library size.} Test accuracy (\%) with DeepSeek-V3 for $K\in\{3,5,10\}$. Entries are individual evaluations; bold marks the highest observed accuracy per task.}
\label{tab:k_sweep}
\small
\setlength{\tabcolsep}{8pt}
\begin{tabular}{l ccc}
\toprule
\textbf{Task} & $K{=}3$ & $K{=}5$ (default) & $K{=}10$ \\
\midrule
\musr{} murder       & \textbf{80.0} & 75.0          & 77.0 \\
\musr{} object       & \textbf{71.7} & 68.9          & 51.9 \\
\musr{} team         & 61.0          & \textbf{68.0} & 62.0 \\
NatPlan meeting      & 49.0          & \textbf{51.0} & 45.0 \\
\bottomrule
\end{tabular}
\end{table}

\begin{figure}[!htbp]
\centering
\includegraphics[width=0.60\linewidth]{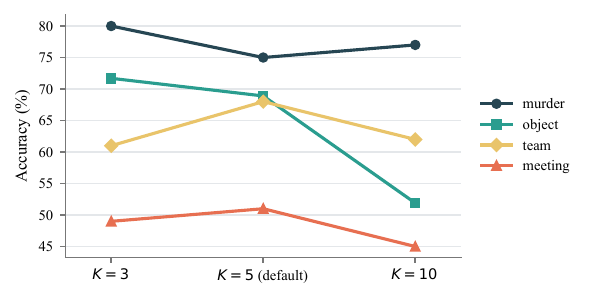}
\caption{\textbf{Accuracy by maximum library size.} DeepSeek-V3 test accuracy for $K\in\{3,5,10\}$; values from Table~\ref{tab:k_sweep}.}
\label{fig:app_ksweep}
\end{figure}

\paragraph{Trace filtering and specification visibility.} Table~\ref{tab:variants} crosses successful-only versus unfiltered training trajectories with whether the agent sees the full agent subroutine specifications. In every condition, the agent sees names, argument schemas, and short descriptions, and a separate LLM call receives the full specification to execute an agent subroutine. \emph{Hidden} therefore means hidden from the agent's prompt, not absent from execution. Correctness filtering improves the hidden-specification condition on all three tasks. Exposing full specifications to the agent does not add a consistent benefit and lowers accuracy on murder and team for the successful-only libraries. This supports the default separation between a short interface for selecting agent subroutines and a full specification for executing them.

\begin{table}[!htbp]
\centering
\caption{\textbf{Trace filtering and specification visibility.} Run-level test accuracy (\%) with DeepSeek-V3 on \musr{}. Visibility refers to the full specification in the agent's prompt; agent subroutine execution calls receive it in all conditions. Variant A is the default. Bold marks the highest accuracy per task.}
\label{tab:variants}
\small
\setlength{\tabcolsep}{8pt}
\begin{tabular}{c cc ccc}
\toprule
\textbf{Variant} & \textbf{Trace filter} & \textbf{Full specification} & \musr{} murder & \musr{} object & \musr{} team \\
\midrule
\textbf{A}  & correct & hidden       & \textbf{75.0} & \textbf{68.9} & 68.0          \\
B           & correct & visible & 69.0          & \textbf{68.9} & 63.0          \\
C           & any     & hidden       & 74.0          & 66.0          & 60.0          \\
D           & any     & visible & 70.0          & 67.0          & \textbf{73.0} \\
\bottomrule
\end{tabular}
\end{table}

\begin{figure}[!htbp]
\centering
\includegraphics[width=0.75\linewidth]{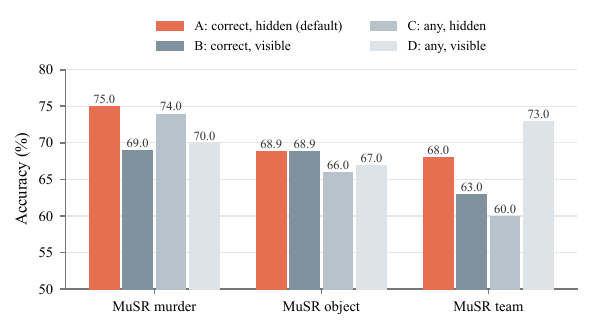}
\caption{\textbf{Trace filtering and specification visibility.} DeepSeek-V3 test accuracy for variants A--D; values from Table~\ref{tab:variants}.}
\label{fig:app_variants}
\end{figure}

\FloatBarrier
\section{Repeated Evaluations on \musr{}}
\label{appendix:multiseed}

Table~\ref{tab:multiseed_musr} reports DeepSeek-V3 means and standard deviations over three runs of each listed method on the same \musr{} test partitions.

\begin{table}[!htbp]
\centering
\caption{\textbf{Repeated DeepSeek-V3 evaluations.} Test accuracy (\%), reported as mean $\pm$ sample standard deviation across three runs on each fixed \musr{} test partition. Bold marks the highest mean per task.}
\label{tab:multiseed_musr}
\small
\setlength{\tabcolsep}{8pt}
\begin{tabular}{l ccc}
\toprule
\textbf{Method}    & \textbf{murder}         & \textbf{object}         & \textbf{team} \\
\midrule
CoT                & $62.3 \pm 2.5$          & $41.5 \pm 2.5$          & $63.3 \pm 2.1$ \\
PoT                & $63.3 \pm 9.1$          & $62.6 \pm 4.6$          & $63.7 \pm 4.6$ \\
ReAct              & $54.3 \pm 8.1$          & $46.9 \pm 3.3$          & $35.7 \pm 2.5$ \\
AWM                & $38.3 \pm 7.6$          & $49.7 \pm 14.1$         & $35.7 \pm 12.5$ \\
\midrule
\textbf{Ours (\method{})}   & $\mathbf{71.0 \pm 4.0}$ & $\mathbf{68.9 \pm 3.8}$ & $\mathbf{68.7 \pm 3.1}$ \\
\bottomrule
\end{tabular}
\end{table}

\begin{figure}[!htbp]
\centering
\includegraphics[width=0.95\linewidth]{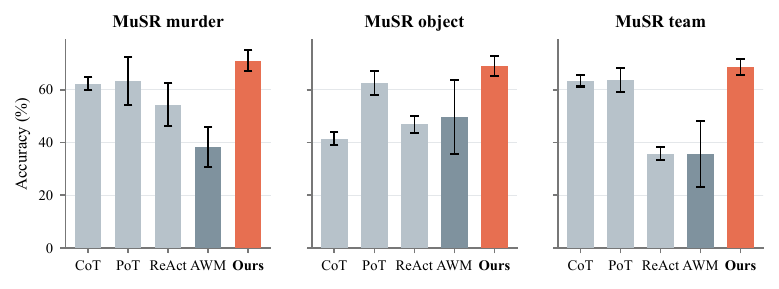}
\caption{\textbf{Repeated DeepSeek-V3 evaluations.} Mean test accuracy with sample standard deviation over three runs; values from Table~\ref{tab:multiseed_musr}.}
\label{fig:app_multiseed}
\end{figure}

\method{} has the highest mean on all three tasks, exceeding PoT by $7.7$, $6.3$, and $5.0$ percentage points on murder, object, and team. Thus the advantage extends beyond the single-run evaluations, including comparisons where the competing methods vary substantially across runs.

\FloatBarrier
\section{Cost Analysis}
\label{appendix:compute}

Induced agent subroutines change how the agent allocates inference: a step in its trajectory can invoke a separate, task-specific LLM computation. Table~\ref{tab:compute_grid} reports total cost, combining offline adaptation and test-time inference. These measurements complement accuracy by describing the computation used to execute the learned procedures.
\begin{table}[!htbp]
\centering
\caption{\textbf{Total cost per task.} US dollars for one offline adaptation plus inference on the full test set. Avg.\ is the unweighted five-task mean. $^\dagger$: served through a different provider (Appendix~\ref{appendix:baselines}).}
\label{tab:compute_grid}
\small
\renewcommand{\arraystretch}{0.92}
\setlength{\tabcolsep}{5.5pt}
\begin{tabular}{l ccccc >{\columncolor{gray!8}}c}
\toprule
& \multicolumn{3}{c}{\musr{}} & RuleArena & NatPlan & Avg. \\
\cmidrule(lr){2-4}\cmidrule(lr){5-5}\cmidrule(lr){6-6}\cmidrule(lr){7-7}
\textbf{Method} & murder & object & team & NBA & meeting & \\
\midrule
\rowcolor{gray!12}\multicolumn{7}{l}{\textsc{DeepSeek-V3}} \\
\midrule
ReAct   & 0.56 & 0.54 & 0.45 & 1.13 & 0.50 & 0.64 \\
\midrule
AWM     & 1.18 & 1.21 & 0.88 & 15.74 & 1.02 & 4.01 \\
ACE     & 3.20\rlap{$^\dagger$} & 2.41\rlap{$^\dagger$} & 2.28\rlap{$^\dagger$} & 12.19\rlap{$^\dagger$} & 1.93\rlap{$^\dagger$} & 4.40\rlap{$^\dagger$} \\
ASI     & 1.81\rlap{$^\dagger$} & 0.92\rlap{$^\dagger$} & 0.57\rlap{$^\dagger$} & 3.02\rlap{$^\dagger$} & 0.45\rlap{$^\dagger$} & 1.35\rlap{$^\dagger$} \\
\midrule
\textbf{Ours (\method{})} & 1.49 & 0.84 & 0.90 & 21.26 & 1.67 & 5.23 \\
\midrule
\rowcolor{gray!12}\multicolumn{7}{l}{\textsc{Gemini 2.5 Flash}} \\
\midrule
ReAct   & 0.36 & 0.38 & 0.75 & 1.91 & 1.76 & 1.03 \\
\midrule
AWM     & 1.06 & 0.69 & 1.59 & 4.41 & 2.42 & 2.03 \\
ACE     & 8.84 & 6.83 & 9.23 & 30.00 & 10.78 & 13.14 \\
ASI     & 2.28 & 1.68 & 3.22 & 4.49 & 3.10 & 2.95 \\
\midrule
\textbf{Ours (\method{})} & 5.28 & 1.28 & 3.78 & 4.25 & 3.55 & 3.63 \\
\bottomrule
\end{tabular}
\end{table}
\FloatBarrier
\section{Paired Comparisons and Response Completion}
\label{appendix:bootstrap}

Paired comparisons help distinguish broad average gains from task-specific behavior. 

\paragraph{Statistical procedure.} We use 95\% percentile-bootstrap intervals with 10\,000 case-level resamples. For repeated evaluations, each case is first averaged within method, then paired differences are resampled. Missing and invalid answers count as incorrect in primary accuracy. The conditional comparison in Table~\ref{tab:families_delta} averages valid-answer repeats only and retains cases with at least one valid answer from each method.

The single-run intervals favor the induced library over ReAct on all tasks except trip, and over PoT on NBA and meeting. In the Gemini sensitivity analysis, gains remain positive on murder, object, NBA, and meeting after restricting to valid-answer repeats, although their magnitude decreases on murder and NBA. The gains therefore persist in comparisons that separate answer completion from accuracy among returned answers. Table~\ref{tab:families_cot_repeat} reports the corresponding repeated comparison with CoT: the intervals exclude zero on object and meeting, include zero on murder and NBA, and show no gain on team.
\begin{table}[!htbp]
\centering
\caption{\textbf{Repeated comparison with CoT on Gemini 2.5 Flash.} Mean test accuracy (\%) over three evaluations of the fixed induced library and three CoT runs on the same test cases; $\Delta$ is the paired induced-minus-CoT difference (percentage points) with a 95\% paired case-bootstrap interval. The induced runs are those of Table~\ref{tab:families_delta}.}
\label{tab:families_cot_repeat}
\footnotesize
\setlength{\tabcolsep}{6pt}
\begin{tabular}{l c c r c}
\toprule
\textbf{Task} & \textbf{Ours (\method{})} & \textbf{CoT} & $\Delta$ (pp) & \textbf{95\% CI} \\
\midrule
\musr{} murder       & 87.3 & 81.3 & $+6.0$  & $(-0.7,\,13.0)$ \\
\musr{} object       & 74.2 & 59.1 & $+15.1$ & $(6.0,\,23.9)$ \\
\musr{} team         & 74.3 & 75.0 & $-0.7$  & $(-6.0,\,4.3)$ \\
RuleArena NBA        & 71.2 & 66.2 & $+5.1$  & $(-3.5,\,13.6)$ \\
Natural Plan meeting & 81.3 & 69.7 & $+11.7$ & $(2.3,\,21.0)$ \\
\bottomrule
\end{tabular}
\end{table}

\begin{figure}[!htbp]
\centering
\includegraphics[width=0.60\linewidth]{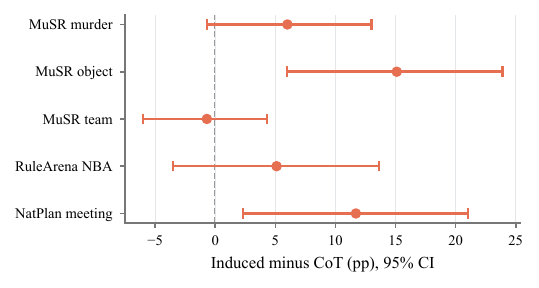}
\caption{\textbf{Repeated comparison with CoT.} Paired induced-minus-CoT difference on Gemini 2.5 Flash with 95\% case-bootstrap intervals; values from Table~\ref{tab:families_cot_repeat}.}
\label{fig:app_cot_repeat}
\end{figure}

\begin{table}[!htbp]
\centering
\caption{\textbf{Sensitivity to invalid or absent answers.} Paired induced-library minus ReAct accuracy differences (percentage points), Gemini 2.5 Flash. All-instance $\Delta$ counts every failure as incorrect; conditional $\Delta$ uses valid-answer repeats only, with $n$ retained cases. Invalid/absent counts are summed over repeats (ReAct / Induced), including out-of-range answer indices. These separate repeat evaluations are not the main-table aggregates.}
\label{tab:families_delta}
\footnotesize
\setlength{\tabcolsep}{4pt}
\begin{tabular}{l l r r r c}
\toprule
\textbf{Model} & \textbf{Task} & All-instance $\Delta$ & Conditional $\Delta$ & $n$ & Invalid/absent R/I \\
\midrule
Flash & murder  & $+10.7$ & $+7.3$ & 100 & 12/0 \\
Flash & object  & $+19.5$ & $+19.7$ & 106 & 0/1 \\
Flash & team    & $-1.7$ & $-4.0$ & 100 & 10/1 \\
Flash & NBA     & $+5.1$ & $+1.5$ & 65 & 10/0 \\
Flash & meeting & $+9.0$ & $+9.0$ & 100 & 1/0 \\
\bottomrule
\end{tabular}
\end{table}

\begin{figure}[!htbp]
\centering
\includegraphics[width=0.60\linewidth]{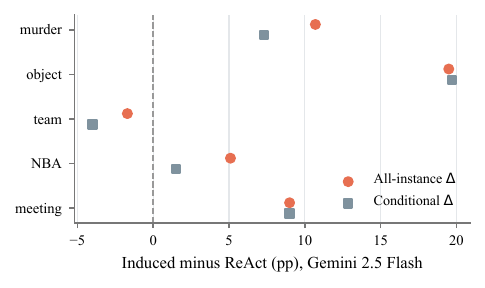}
\caption{\textbf{Sensitivity to invalid or absent answers.} All-instance and conditional induced-minus-ReAct differences on Gemini 2.5 Flash; values from Table~\ref{tab:families_delta}.}
\label{fig:app_delta}
\end{figure}

\begin{table}[!htbp]
\centering
\caption{\textbf{DeepSeek-V3 accuracy and uncertainty.} Test accuracy (\%) with 95\% case-bootstrap intervals in brackets; bold marks the highest point estimate per task. These individual evaluations differ from the repeated-run summaries. Meeting and trip use the deterministic verifier. Intervals condition on the observed test outcomes.}
\label{tab:bootstrap_marginal}
\footnotesize
\setlength{\tabcolsep}{6pt}
\renewcommand{\arraystretch}{1.15}
\begin{tabular}{lccccc}
\toprule
Task & CoT & PoT & ReAct & AWM & Ours (\method{}) \\
\midrule
\musr{} murder & \shortstack{60.0\\$[50.0, 69.0]$} & \shortstack{67.0\\$[58.0, 76.0]$} & \shortstack{45.0\\$[35.0, 55.0]$} & \shortstack{47.0\\$[38.0, 57.0]$} & \shortstack{\textbf{75.0}\\$[66.0, 83.0]$} \\
\addlinespace
\musr{} object & \shortstack{39.6\\$[30.2, 49.1]$} & \shortstack{60.4\\$[50.9, 69.8]$} & \shortstack{50.0\\$[40.6, 59.4]$} & \shortstack{53.8\\$[44.3, 63.2]$} & \shortstack{\textbf{68.9}\\$[60.4, 77.4]$} \\
\addlinespace
\musr{} team & \shortstack{65.0\\$[55.0, 74.0]$} & \shortstack{\textbf{69.0}\\$[60.0, 78.0]$} & \shortstack{38.0\\$[29.0, 48.0]$} & \shortstack{36.0\\$[27.0, 45.0]$} & \shortstack{68.0\\$[59.0, 77.0]$} \\
\addlinespace
RuleArena NBA & \shortstack{48.5\\$[36.4, 60.6]$} & \shortstack{40.9\\$[28.8, 53.0]$} & \shortstack{28.8\\$[18.2, 39.4]$} & \shortstack{59.1\\$[47.0, 71.2]$} & \shortstack{\textbf{74.2}\\$[63.6, 84.8]$} \\
\addlinespace
NatPlan meeting & \shortstack{35.0\\$[26.0, 45.0]$} & \shortstack{27.0\\$[18.0, 36.0]$} & \shortstack{32.0\\$[23.0, 41.0]$} & \shortstack{43.0\\$[33.0, 53.0]$} & \shortstack{\textbf{51.0}\\$[41.0, 61.0]$} \\
\addlinespace
NatPlan trip & \shortstack{0.0\\$[0.0, 0.0]$} & \shortstack{0.0\\$[0.0, 0.0]$} & \shortstack{1.0\\$[0.0, 3.1]$} & \shortstack{\textbf{2.1}\\$[0.0, 5.2]$} & \shortstack{0.0\\$[0.0, 0.0]$} \\
\bottomrule
\end{tabular}
\end{table}

\begin{figure}[!htbp]
\centering
\includegraphics[width=0.95\linewidth]{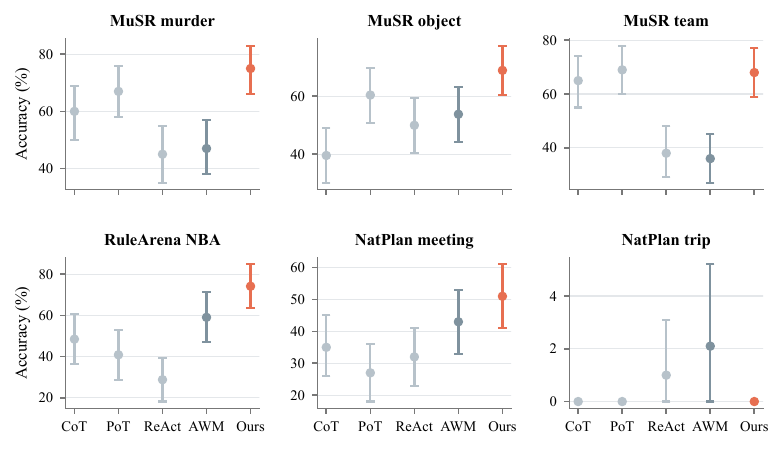}
\caption{\textbf{DeepSeek-V3 accuracy and uncertainty.} Test accuracy with 95\% case-bootstrap intervals; values from Table~\ref{tab:bootstrap_marginal}.}
\label{fig:app_marginal}
\end{figure}

\begin{table}[!htbp]
\centering
\caption{\textbf{Paired single-run differences with DeepSeek-V3.} Induced-library minus baseline accuracy (percentage points), with 95\% paired case-bootstrap intervals for the runs in Table~\ref{tab:bootstrap_marginal}. A star marks an interval excluding zero. Intervals are unadjusted for multiple comparisons.}
\label{tab:bootstrap_paired}
\footnotesize
\setlength{\tabcolsep}{6pt}
\renewcommand{\arraystretch}{1.15}
\begin{tabular}{lcccc}
\toprule
Task & CoT & PoT & ReAct & AWM \\
\midrule
\musr{} murder & \shortstack{$+15.0^{\star}$\\$[+4.0,+26.0]$} & \shortstack{$+8.0$\\$[-1.0,+18.0]$} & \shortstack{$+30.0^{\star}$\\$[+18.0,+41.0]$} & \shortstack{$+28.0^{\star}$\\$[+16.0,+40.0]$} \\
\addlinespace
\musr{} object & \shortstack{$+29.2^{\star}$\\$[+16.0,+41.5]$} & \shortstack{$+8.5$\\$[-0.9,+18.9]$} & \shortstack{$+18.9^{\star}$\\$[+6.6,+31.1]$} & \shortstack{$+15.1^{\star}$\\$[+2.8,+26.4]$} \\
\addlinespace
\musr{} team & \shortstack{$+3.0$\\$[-7.0,+13.0]$} & \shortstack{$-1.0$\\$[-12.0,+10.0]$} & \shortstack{$+30.0^{\star}$\\$[+17.0,+43.0]$} & \shortstack{$+32.0^{\star}$\\$[+19.0,+44.0]$} \\
\addlinespace
RuleArena NBA & \shortstack{$+25.8^{\star}$\\$[+12.1,+39.4]$} & \shortstack{$+33.3^{\star}$\\$[+19.7,+47.0]$} & \shortstack{$+45.5^{\star}$\\$[+28.8,+62.1]$} & \shortstack{$+15.2$\\$[0.0,+31.8]$} \\
\addlinespace
NatPlan meeting & \shortstack{$+16.0^{\star}$\\$[+7.0,+25.0]$} & \shortstack{$+24.0^{\star}$\\$[+14.0,+34.0]$} & \shortstack{$+19.0^{\star}$\\$[+9.0,+29.0]$} & \shortstack{$+8.0$\\$[-1.0,+17.0]$} \\
\addlinespace
NatPlan trip & \shortstack{$0.0$\\$[0.0,0.0]$} & \shortstack{$0.0$\\$[0.0,0.0]$} & \shortstack{$-1.0$\\$[-3.1,0.0]$} & \shortstack{$-2.1$\\$[-5.2,0.0]$} \\
\bottomrule
\end{tabular}
\end{table}

\begin{figure}[!htbp]
\centering
\includegraphics[width=0.95\linewidth]{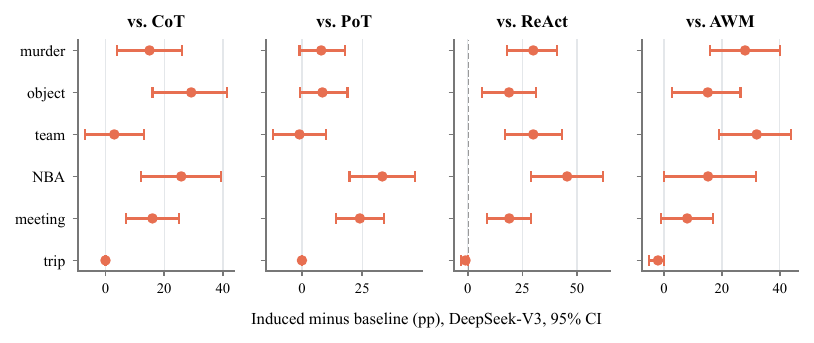}
\caption{\textbf{Paired single-run differences with DeepSeek-V3.} Induced-library minus baseline accuracy with 95\% paired case-bootstrap intervals; values from Table~\ref{tab:bootstrap_paired}.}
\label{fig:app_paired}
\end{figure}

\FloatBarrier
\subsection{Cross-model transfer statistics}
\label{appendix:transfer_details}

Table~\ref{tab:transfer_details} reports the paired accuracy gains and confidence intervals for Table~\ref{tab:transfer}. Libraries induced by Gemini 2.5 Flash are deployed unchanged on Gemini 3.5 Flash-Lite and compared with that target model's ReAct agent on the same test instances. Each difference is formed from repeated measurements of the same cases, before rounding the accuracies. All five intervals exclude zero, supporting reuse of the learned procedures across executing models.

\begin{table}[!htbp]
\centering
\caption{\textbf{Paired cross-model transfer gains.} $\Delta$ is the transferred-library accuracy minus target-model ReAct accuracy, in percentage points, with the corresponding 95\% confidence interval. The source model is Gemini 2.5 Flash and the target model is Gemini 3.5 Flash-Lite.}
\label{tab:transfer_details}
\small
\setlength{\tabcolsep}{6pt}
\begin{tabular}{l r c}
\toprule
\textbf{Task} & $\Delta$ (pp) & \textbf{95\% CI} \\
\midrule
\musr{} murder       & $+14.7$ & $(8.3,\,21.3)$ \\
\musr{} object       & $+12.3$ & $(3.5,\,21.4)$ \\
\musr{} team         & $+10.7$ & $(3.2,\,18.5)$ \\
RuleArena NBA        & $+32.8$ & $(22.7,\,42.9)$ \\
Natural Plan meeting & $+12.3$ & $(6.3,\,18.3)$ \\
\bottomrule
\end{tabular}
\end{table}

\begin{figure}[!htbp]
\centering
\includegraphics[width=0.60\linewidth]{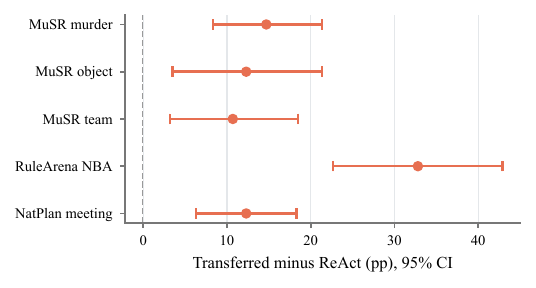}
\caption{\textbf{Paired cross-model transfer gains.} Transferred-library minus target-model ReAct accuracy with 95\% confidence intervals; values from Table~\ref{tab:transfer_details}.}
\label{fig:app_transfer_ci}
\end{figure}

\FloatBarrier
\section{Additional-Sampling Control}
\label{appendix:sc20}

We evaluate self-consistency~\citep{wang2023selfconsistency} with 20 CoT samples per instance on DeepSeek-V3. Samples use $T{=}0.7$ with response caching disabled. The final answer is the majority of successfully parsed answer indices, with ties resolved by the earliest vote. If no sample yields a valid answer index, the instance is scored as incorrect. This tests whether additional sampling without a learned library, a common way to scale test-time compute~\citep{snell2025scaling}, offers a competitive alternative.

\begin{table}[!htbp]
\centering
\caption{\textbf{Self-consistency with 20 samples.} DeepSeek-V3 test accuracy (\%) on the same fixed test cases as the induced-library runs in Table~\ref{tab:bootstrap_marginal}. $\Delta$ is SC@20 minus induced-library accuracy, in percentage points.}
\label{tab:sc20}
\small
\setlength{\tabcolsep}{10pt}
\begin{tabular}{l c c c}
\toprule
\textbf{Task} & \textbf{SC$@N{=}20$} & \textbf{Ours (\method{})} & $\Delta$ \\
\midrule
\musr{} murder & 66.0 & 75.0 & $-9.0$  \\
\musr{} object & 45.3 & 68.9 & $-23.6$ \\
\bottomrule
\end{tabular}
\end{table}

\begin{figure}[!htbp]
\centering
\includegraphics[width=0.50\linewidth]{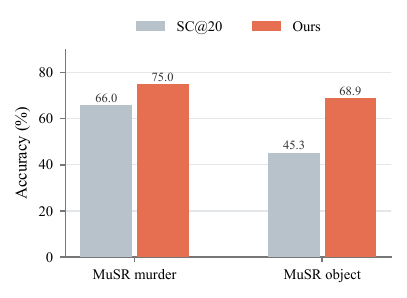}
\caption{\textbf{Self-consistency with 20 samples.} DeepSeek-V3 test accuracy of SC@20 and the induced library; values from Table~\ref{tab:sc20}.}
\label{fig:app_sc20}
\end{figure}

SC@20 remains below the induced-library agent by 9.0 points on murder and 23.6 points on object. The comparison supports learning reusable agent subroutines as an alternative to repeatedly sampling complete solutions.

\FloatBarrier
\section{Tasks Outside the Main Evaluation}
\label{appendix:out_of_scope}
\label{appendix:oos}

Our main evaluation studies whether recurring reasoning procedures can be learned from successful agent experience and reused on new instances. Supplementary evaluations on other RuleArena and Natural Plan subtasks highlight two considerations for extending this approach: obtaining sufficient successful experience for induction and choosing suitable implementations for the learned procedures.

\paragraph{Insufficient successful experience.} On Natural Plan trip and calendar, the collection agent produces 2 and 19 successful trajectories, too few to cover the tasks' recurring planning operations. On trip, accuracy remains near zero for the induced libraries and for every evaluated method (Table~\ref{tab:bootstrap_marginal}). Better experience collection would extend induction to such tasks.

\paragraph{Calculation-heavy tasks.} RuleArena airline (baggage-fee calculation, e.g., applying route- and cabin-specific allowances and overweight and oversize surcharges to several bags on a Mumbai--Minneapolis flight) and tax (tax calculation, e.g., determining the balance owed by a married couple filing jointly after combining wage and retirement income, applying deductions and credits, and subtracting withheld tax) require long numerical calculations under strict rules, which suit code execution: a Python program induced for these subtasks extracts over 20 parameters from the problem statement and computes the answer directly. Their rule texts are also long, which makes agent training costly (about \$800 per training run). We therefore evaluate arithmetic-heavy reasoning on AIME instead.

\end{document}